\documentclass[final]{cvpr}

\usepackage{times}
\usepackage{epsfig}
\usepackage{amsmath}
\usepackage{amssymb}

\usepackage{url}            
\usepackage{booktabs}       
\usepackage{amsfonts}       
\usepackage{nicefrac}       
\usepackage{microtype}      
\usepackage{pifont}
\usepackage{caption}
\usepackage{wrapfig}
\usepackage{subcaption}

\ifx\theorem\undefined
\newtheorem{theorem}{Theorem}
\fi
\ifx\assumption\undefined
\newtheorem{assumption}[theorem]{Assumption}
\fi
\ifx\condition\undefined
\newtheorem{condition}{Condition}
\fi
\ifx\example\undefined
\newtheorem{example}{Example}
\fi
\ifx\property\undefined
\newtheorem{property}{Property}
\fi
\ifx\lemma\undefined
\newtheorem{lemma}[theorem]{Lemma}
\fi
\ifx\proposition\undefined
\newtheorem{proposition}[theorem]{Proposition}
\fi
\ifx\remark\undefined
\newtheorem{remark}{Remark}
\fi
\ifx\corollary\undefined
\newtheorem{corollary}[theorem]{Corollary}
\fi
\ifx\definition\undefined
\newtheorem{definition}[theorem]{Definition}
\fi
\ifx\conjecture\undefined
\newtheorem{conjecture}[theorem]{Conjecture}
\fi
\ifx\fact\undefined
\newtheorem{fact}[theorem]{Fact}
\fi
\ifx\claim\undefined
\newtheorem{claim}[theorem]{Claim}
\fi

\newcommand{\xmark}{N}%

\newcommand{\jmean}{\mathcal J_{\mathcal M}}
\newcommand{\jrecall}{\mathcal J_{\mathcal O}}
\newcommand{\jdecay}{\mathcal J_{\mathcal D}}
\newcommand{\fmean}{\mathcal F_{\mathcal M}}
\newcommand{\frecall}{\mathcal F_{\mathcal O}}
\newcommand{\fdecay}{\mathcal F_{\mathcal D}}

\usepackage[pagebackref=true,breaklinks=true,colorlinks,bookmarks=false]{hyperref}

\begin{document}

\title{  Convolutional Networks as Extremely Small Foundation Models: \\
Visual Prompting and Theoretical Perspective}

\author{Jianqiao Wangni\\
Tsinghua University\\
Beijing, China\\
{\tt\small zjnqha@gmail.com}
}

\maketitle

\begin{abstract}
Comparing to deep neural networks trained for specific tasks, those foundational deep networks trained on generic datasets such as ImageNet classification, benefits from larger-scale datasets, simpler network structure and easier training techniques. In this paper, we design a prompting module which performs few-shot adaptation of generic deep networks to new tasks. Driven by learning theory, prompting modules are as simple as possible, as they generalize better under the same training error. We use a case study on video object segmentation to experiment, with a concrete prompting implementation, the Semi-parametric Deep Forest (SDForest) that combines several nonparametric methods such as correlation filter, random forest, image-guided filter, with a deep network trained for ImageNet classification task. From a learning-theoretical point of view, all these models are of significantly smaller VC dimension or complexity so tend to generalize better, as long as the empirical studies show that the training error of this simple ensemble can achieve comparable results from a end-to-end trained deep network. We also propose a novel methods of analyzing the generalization under the setting of video object segmentation to make the bound tighter. In practice, SDForest has extremely low computation cost and achieves real-time even on CPU. We test on video object segmentation tasks and achieve competitive performance at DAVIS2016 and DAVIS2017  with purely deep learning approaches, without any training or fine-tuning. 
\end{abstract}

\section{Introduction}
Occam's razor principle states that \textit{plurality should not be posited without necessity}, which is a precursor for rich literature that supports simple math and physics explanation behind the complex phenomenon, as well as later development of learning theory that favors simple models, which are academically measured by VC dimension or Radamecar complexity \cite{mohri2018foundations}. Suppose that two linear regression models reach the same training error on the same task, then the simpler one, or the one with fewer parameters, would perform better on real-world data, or we put it as low model complexity achieves smaller generalization error. In this paper, we try to study the model complexity versus generalization error trade-off on the problem of object segmentation, which has been a long-standing task in computer vision. The objective of such a task is to classify each pixel into various objects and the background. It can be achieved by clustering algorithms, although most deep learning-based algorithms take the problem as pixel-wise classification. 
\begin{figure*}[t]
	\begin{center}
		\includegraphics[width=\textwidth]{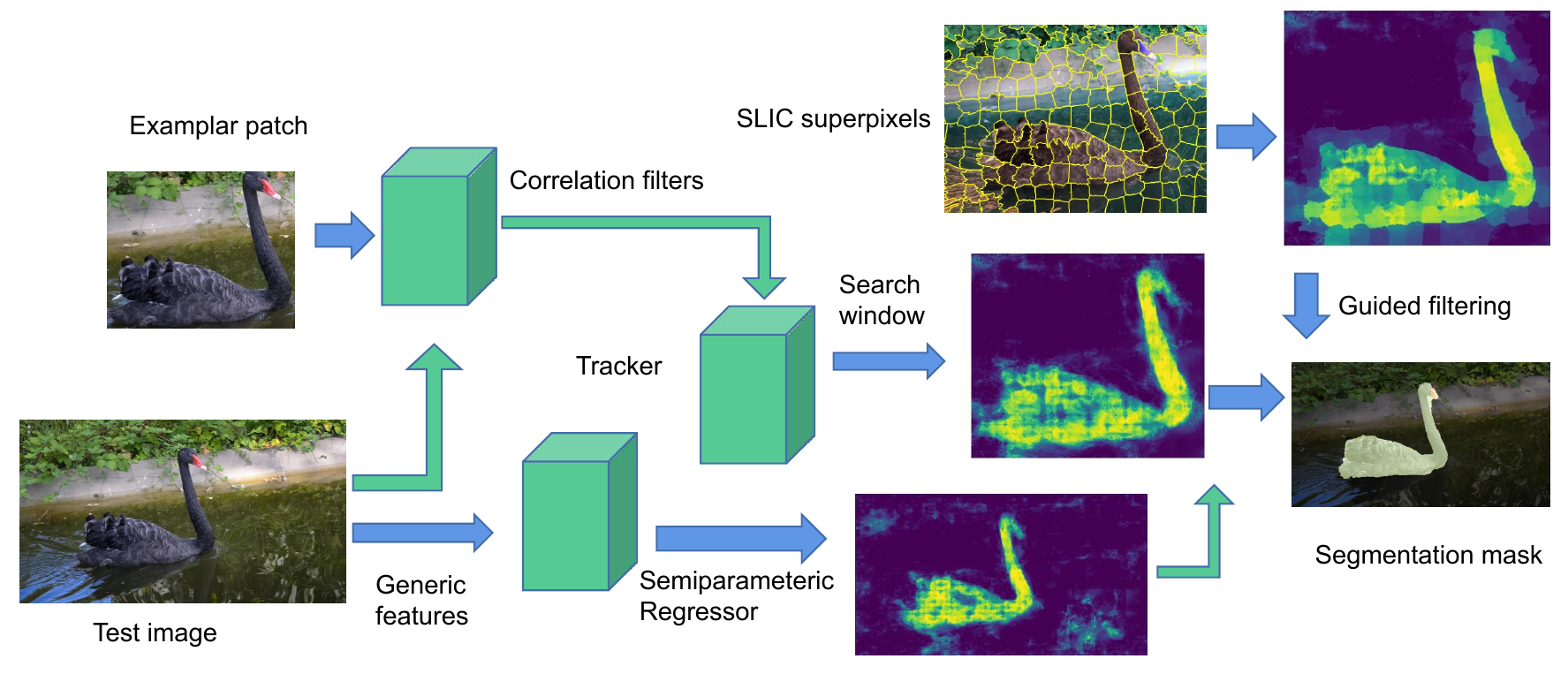}
	\end{center}
	\caption{Workflow of SDForest for video object segmentation.}\label{fig:norm}
	\vspace{-10pt}
	\label{fig:workflow}
\end{figure*}

We understand the trend of deep learning approaches to computer vision mainly benefiting from several directions: larger-scale datasets, over-parameterization and end-to-end learning. Over-parameterization refers to a large number of network weights that outnumber the training images.  End-to-end training describes a paradigm that the algorithm consists of a sequence of differentiable math operators, so that model is trained through back-propagation, and for testing images, the prediction is generated by forward-propagation along with the operators.

Human civilization stepped into GPT era by \textit{fine-tuning and 
 prompting} a pretrained large language models. In computer vision, video object segmentation resembles the tasks to prompting: the users show the object boundary in the first frame and let the deep models to memorize it soon enough shortly before handling the following frames. 
 
There is the thought about \textit{simple prompting on a complex foundation model}.  
The \textit{ subjective} concepts, like cats, dogs, bikes and cars are defined by humans and there lacks math or physics language to characterize, provide the deep learning approaches a more successful stage to perform. 
But for \textit{objective} tasks  like segmentation, the concepts of boundary, corners, and contours are mathematically meaningful using the language of the gradient, Laplacian operators, and eigenvalues \cite{shi1994good, lowe2004distinctive, bay2006surf}. 
These hand-craft features are easily \textit{promptable}, as they learn these targeted visual features with very few examples, so it suits the robotic tasks, localization and mapping. A typical counter-example of learning objective concept would be: imagine a powerful ResNet learning to solve a quadratic equation through fitting one billion groups of coefficients and their corresponding roots, is inefficient, economically impractical and mathematically inaccurate, while a simple formula $x=({-b \pm \sqrt{b^2-4ac})/(2a) }$ gives a closed-form solution. Of course, this example goes to another extreme, but it sheds some inspiration that known physics or math principles should not be encoded to any learning algorithm on whatever datasets.

Admittedly that training deep networks for a specific task may be the best fit  by now, successfully enough to somehow resemble human intelligence, as the empirical generalization error are within our acceptance of such problems, e.g. we consider a neural network recognizing $\%60$ of ImageNet objects to be intelligent, but such an error is far from impractical for object tasks like segmentation or computational photography. 
These tasks generally require label consistency among neighboring pixels. We could fine-tune a deep model with conditional random fields (CRF)  \cite{chen2017deeplab} as header, to incorporate inter-pixel constraints, which essentially redistribute or reweight the loss function each pixel depending on classes of their neighbors. This simply adjusts the trade-off of pixel-wise prediction, depending on the potential function and training set, but not on testing images, and whether this boosts the performance would depend on distribution shift between the testing images and the training images. 
Rigorously, even if the training error is zero and all pairwise pixel relations satisfy the property, this still does not transpose to an arbitrary new image, due to the generalization error, part of which is linearly related to the VC dimension of the model. 
Currently, the VC dimension of deep neural networks is still not clear enough to provide a satisfactory generalization error to meet the empirical observation \cite{bartlett2019nearly}. 



In this paper, we are motivated by the the success of zero-shot and few-shot prompting of pretrained large models, and try to create to learning-free methods as possible, to create a methodology of only involving testing data. We tend to 
transfer those feature extractors from unrelated tasks could boost the performance over handcrafted features, and by comparison, having no way to overfit anything as having no prior information about this task whatsoever. 

\begin{figure*}[t]
\begin{subfigure}{.5\textwidth}
  \centering
  \includegraphics[width=\textwidth]{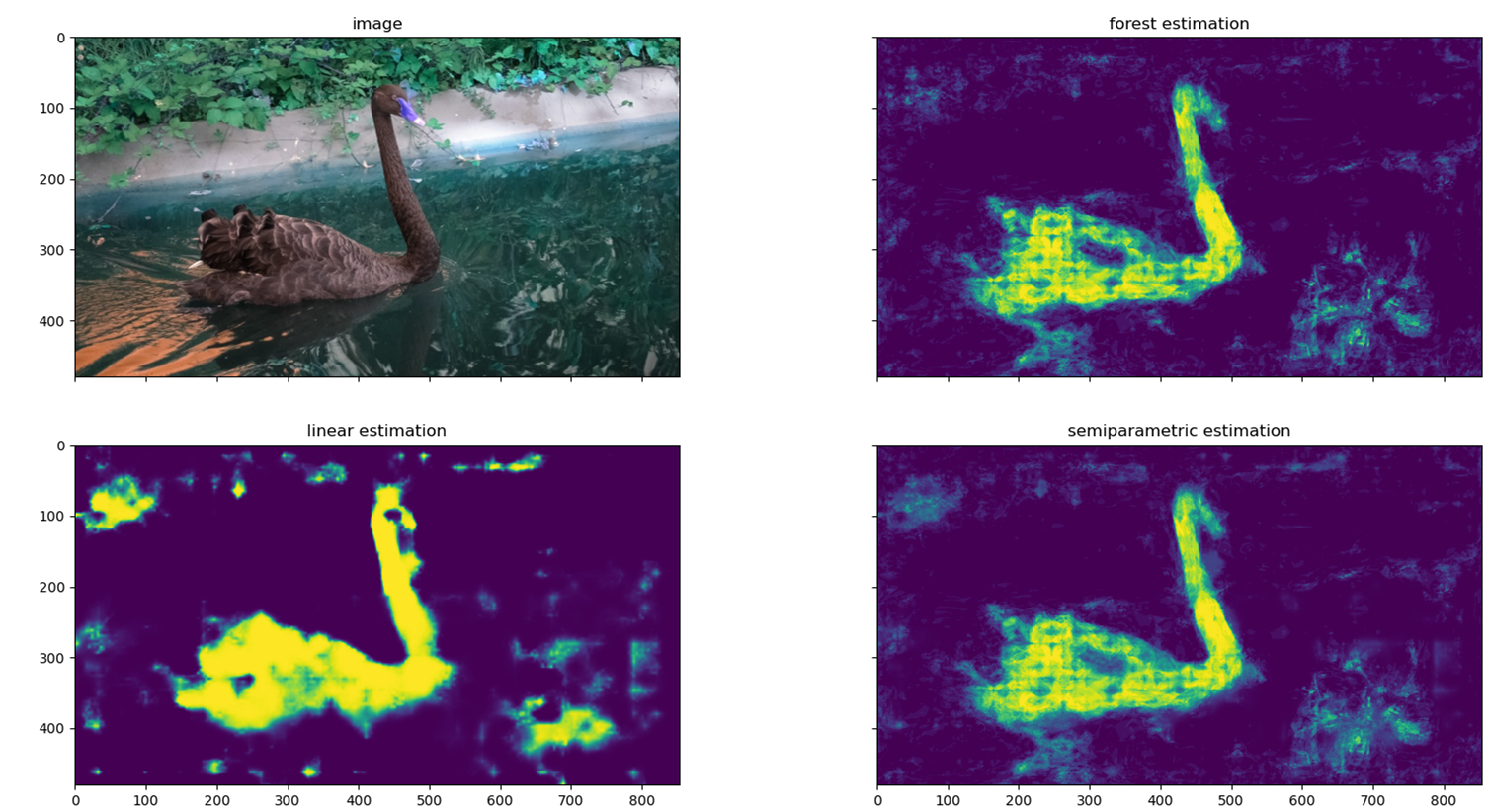}
  \caption{Davis2016 example: Blackswan}
  \label{fig:sfig1}
\end{subfigure}%
\begin{subfigure}{.5\textwidth}
  \centering
  \includegraphics[width=\textwidth]{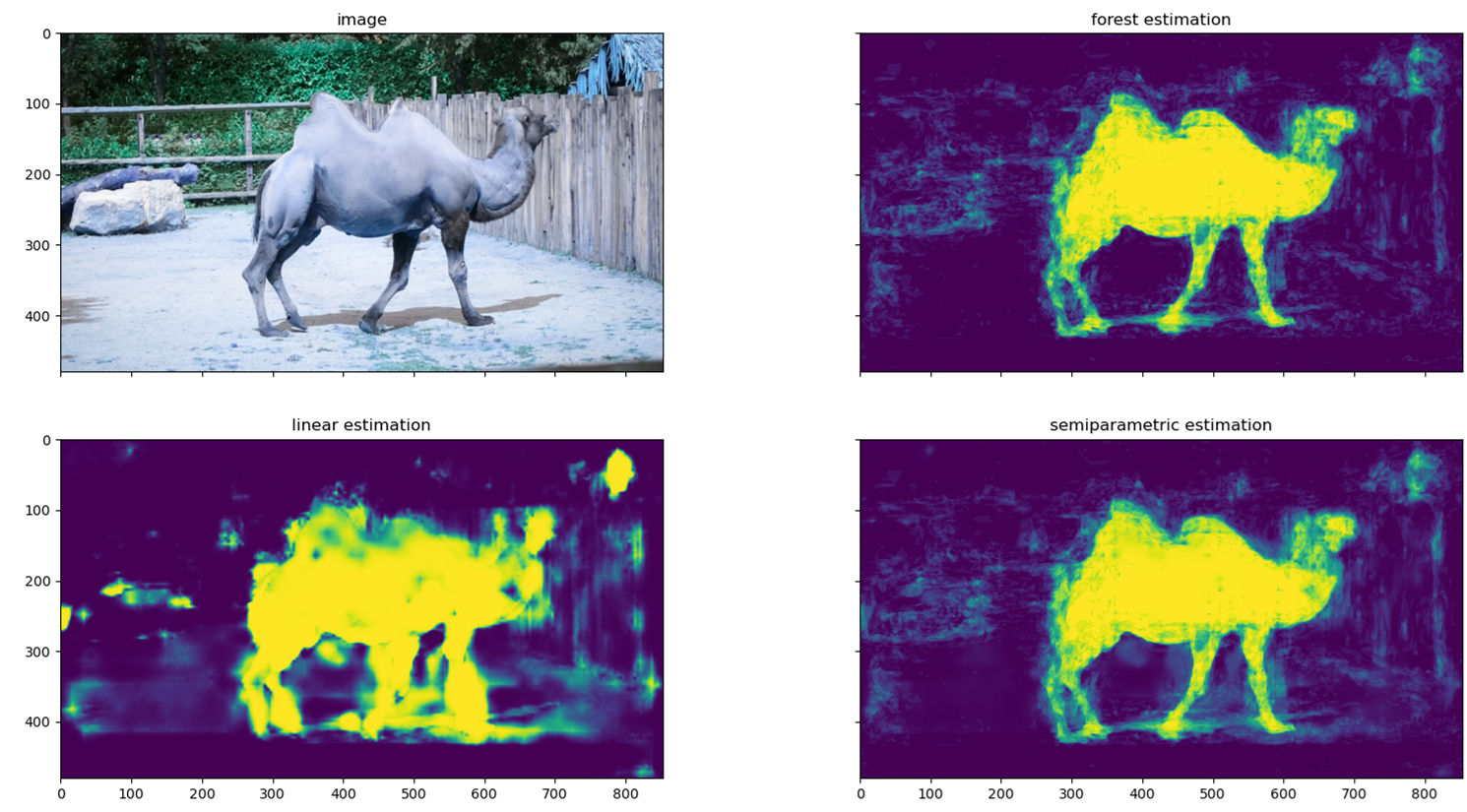}
  \caption{DAVIS 2016 example: camel}
  \label{fig:sfig2}
\end{subfigure}
\caption{Visualization of semi-parametric estimation for VOS. From left to right, and from top to bottom: original image, random forest, linear model on deep features, and semi-parametric ensemble.} 
\vspace{-10pt}
\label{fig:fig}
\end{figure*}

\section{Background}
This paper aims to provide a prompting module for video object segmentation (VOS) with a foundational deep model. The prompting module uses simple parametric and nonparametric algorithms, to memorize the first frame of the video has a ground-truth mask provided as reference, to which is generally referred as semi-supervised VOS. We will first introduce a deep learning baseline method that is end-to-end trained, which we will compare against. We also benefit from its tracking part to narrow down a search window for our algorithm.

We cannot enumerate most of advancements in video object segmentation: researchers proposed different backbone network architecture and methods of aggregating multiscale features, both locally and globally to the mask prediction, including Fully Convolutional Network(FCN) \cite{long2015fully}, Pyramid Scene Parsing Network (PSPnet) \cite{zhao2017pyramid}, DeepLab \cite{chen2017deeplab,chen2017rethinking} and U-Net \cite{ronneberger2015u}. These architecture innovations make neural networks possibly more expressive and more capable to fit, which, very much resembles the effect of residual connection make a sequential network more expressive. The end-to-end learning methods witnessed great advancement in the area of object detection and instance segmentation, e.g., Mask-RCNN \cite{he2017mask} used the region proposal network (RPN) to localize a Region of Interest (ROI) and perform segmentation on ROI individually, 
and the training would be much difficult if ROI is generated by selective search instead of RPN. Besides, the attention mechanism that assigns different weight to different locations of the grid 
\cite{chen2016attention, fu2019dual} and recurrent network architecture 
\cite{visin2016reseg} 
also coarsely fall into these two categories. 
Both directions seem to be the opposite of the simplicity.

SiamMask \cite{wang2019fast}  is a framework that unifies both tracking and segmentation in one deep network. One branch of the predictor has a MaskRCNN header \cite{he2017mask} to predict the mask and it uses a convolutional Siamese network \cite{bertinetto2016fully} to measure the similarity of a given patch and a typically larger search region. Such a framework can generate a response score map for each location densely. We denote the examplar patch as $z$ and denote the search region as $I$, which we assume to be of size $w\times h$. Both patches are fed into a pre-trained convolutional neural network to obtain two deep feature maps $\psi(z)$ and $\psi(I)$. Two maps are then correlated to obtain the response map $\psi(z) \star \psi(I)$. 
Such a convolution reflects the response to candidate windows (RoW) centered at each location. This idea is the support of correlation filter trackers \cite{henriques2014high}. We further denote the $n$-th frame as $I_n$ and the examplar from $(n-1)$th frame as $z_{n-1}$.  On top of these feature maps. SiamMask also predicts the segmentation mask by:
\begin{align}
    \hat y =\mu(\psi(z) \star \psi(I)) \in \mathbb [0,1]^{w \times h},
\end{align}
and the prediction $\mu(I_n)$ represents the confidence map of a pixel being the target object at the $n$-th frame. To efficiently unify segmentation and tracking, each candidate window has two kinds of label to fit, one binary label to indicate whether the whole window contains the object and a mask $y_n \in \{+1,-1\}^{w \times h}$ to indicate pixel-wise confidence. During training, the segmentation prediction branch should take the binary label into account by considering weights $y_n$. We use $i$ as the 2D spatial index, then the loss function for training the mask prediction branch is as follows:
\begin{align*}
\mathcal L(\psi,\mu,I,y)=  \frac{1}{2wh}\sum_{i}\log(1+ \exp({-y[{i}]\mu(I)[i]} ).
\end{align*}
The tasks for multi-object segmentation require classifying each pixel into specific classes. The prediction $\mu(I) \in \mathbb R^{c\times w \times h}$ of an image $I$ should be a $3$-dimensional confidence tensor, each channel indicating the confidence of one class out of $(c-1)$ in total or the background.
\section{Approach}
This section describes the main component of the prompting module, the semi-parametric random forest estimation.  As mentioned above, we will only use some deep networks trained for generic tasks like ImageNet classification. 
From a practical side, in our case of VOS or tracking, the data comes in a stream, so the network may incur a large computation cost to update in real-time accordingly, and there is only the first frame for us to learn. For the scenarios when only a small number of reliable data available, non-parametric methods like nearest neighbors \cite{boiman2008defense} are strong tools in comparison to those parametric models that need iterative training.  However, the computation for the rest frames is expensive for a direct application of nearest neighbor, considering the tens of thousands of pixels as references and the same order of pixels in the following frames, each of which has hundreds of dimensions. The fast query of nearest neighbors can be achieved by KD-trees \cite{bentley1975multidimensional}  also inspires us to use a tree structure to integrate features of pixels. 
From another perspective, as we know, some semantically meaningful segments are mostly in an irregular shape, which brings us difficulty to cluster them with deep features. 
A tree structure with proper depth and number of leaf nodes has a side effect of implicitly grouping nearby pixels into a concrete region, since the leaf node of each tree is sufficient to represent a hypercube in feature space, being somehow coarse-grain to include pixels within each segment, but not so over-complicated as to overfit particular pixels. The ensemble of these models with enough diversity is practically efficient and statistically stable, given prior research in random forests both in theory and implementation.

\begin{figure}[htb]
\centering
{\includegraphics[width=0.5\textwidth]{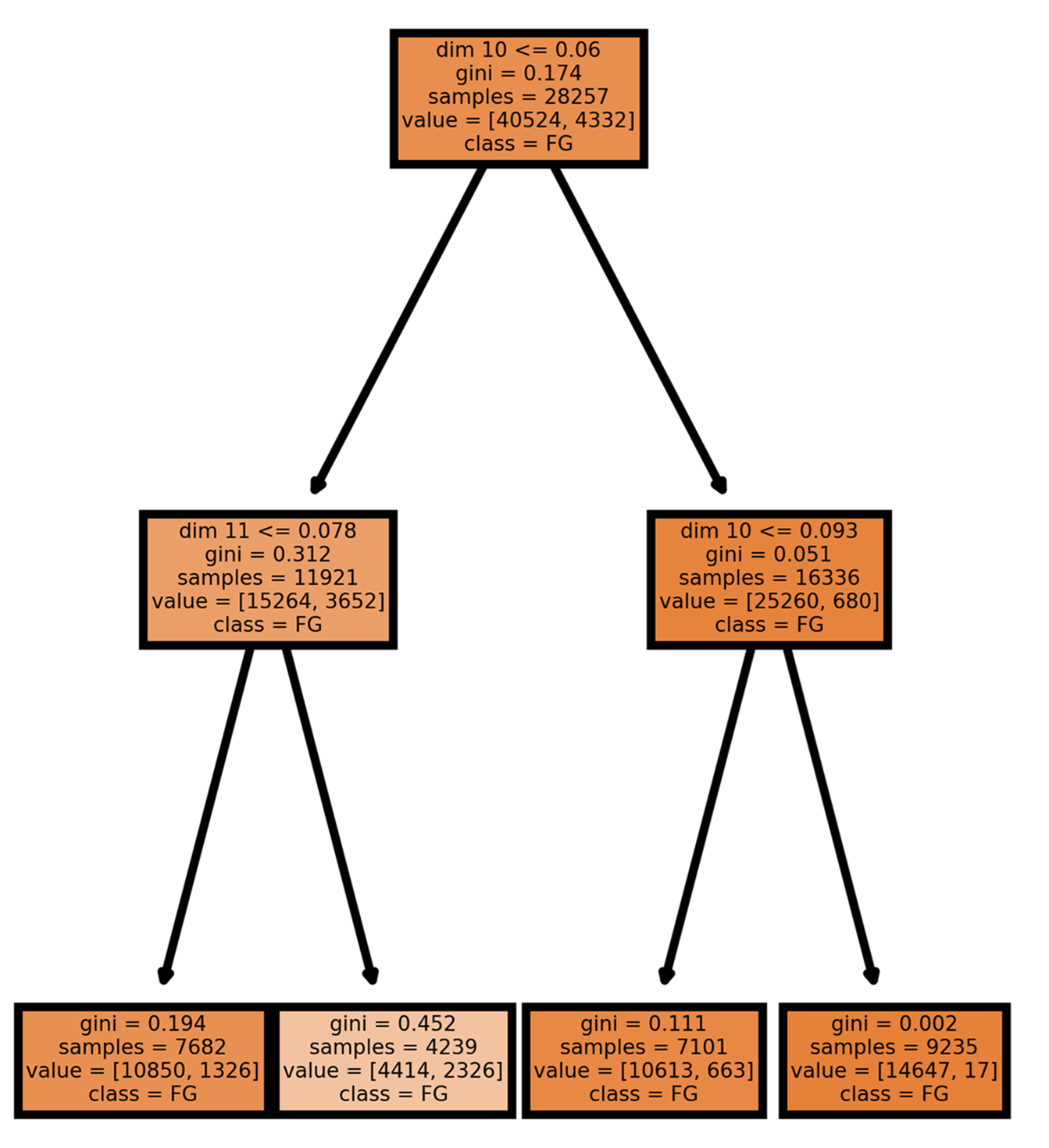}}
\vspace{-10pt}
\caption{ Examples of tree splits within a trained random forest regressor. 
}
\vspace{-10pt}
\label{fig:tree}
\end{figure}
Two standpoints above lead us to a semi-parametric regression model, which is obtained by a random forest classifier and a softmax regression model, the latter of which could refer to the header of deep networks. 
We denote the generic features to be $h(I)$ extracted from an input image $I$. 
Let $\pi_{q}$ represents the feature space that falls within the leaf node $q$, out of $Q$ nodes in total, and let $\omega_m$ be the value of  objectiveness assigned to leaves, then the prediction confidence of $I$ at $i$th pixel is calculated as
\begin{align*}
    P(I)[i] =  \mu(h(I)[i]) +\frac{\gamma}{Q} \sum_{q=1}^{Q} \mathbb E\left\{ \omega_{q} \mathbb I(h(I)[i] \in \pi_{q} )\right\}
\end{align*}
where $\epsilon$ is a threshold and $\gamma$ is the weight of forest estimator, and $\mathbb I$ is the indicator function, and the expectation is taken over all trees.  The parametric linear estimator $\mu$ is trained on the first frame as well, as a task of pixel-wise classification into each object or foreground, and we use the confidence value of the estimator as the output.  
The worst-case being, $h$ is not informative and behaves like a random projection, but the semi-parametric estimator would still work: as long as some correct labels in a testing image could exist, the estimator will try to extract patterns from corresponding features and expands the decision criterion to the rest image. We will refer to the estimator as Semi-parametric Deep Forest (SDForest).

\textbf{A learning-theoretical perspective.} The behavior of decision trees approximates nearest neighbors when the trees are deep enough to include every pixel correctly in one leaf node, and finding nearest neighbors is an ideal nonparametric algorithm for predicting a pixel in this context. Learning theory is hereby to give a worst-case guarantee about generalization \cite{shalev2014understanding}. Given $m$ pixels $S=\{x_i=h(I)[i]\}_{i=1}^m$ from a distribution in features space $\mathcal X=\{0,1\}^J$, a decision tree with $Q$ nodes and the splitting rules are set to $\mathbb I(x_d\leq \epsilon_j),j\in [J]$ where $\epsilon_j$ is a threshold, a set of hypotheses $\mathcal F$ corresponding to leaves, then 
with probability of at least $1-\delta$, $  \forall f \in \mathcal F$, 
\begin{align*}
 \mathcal L_D(f) \leq \mathcal L_S(f) +\sqrt{\frac{(n+1) \log (J+3)+ \log (2/\delta)}{2m}},  
\end{align*}
where $\mathcal L_S$ is the empirical loss on the reference frame and the $\mathcal L_D$ is the loss in expectation. In comparison, an almost-tight lower-bound of the VC dimension of deep ReLU networks  \cite{bartlett2019nearly} shows
\begin{align}\label{vc-dl}
\mathcal L_D(f) \leq \mathcal L_S(f) +\sqrt{\frac{VC(\mathcal F)+ \log (2/\delta)}{2m}},\\ VC(\mathcal F) = \Omega \left(\#W \#U\log (\frac{\#W}{\#U})\right),
\end{align}
where $W$ represents network weights, $U$ represents all its layers,  $\#$ is an operator that counts numbers, and typically $m=\Omega(wh)$. To give an intuitive expression with our experimentation setting: using ResNet-50 as the header for segmentation involves an extra 23 million parameters and 50 layers, leads to a VC dimension dominates that of any decision trees on typically about $10,000$ pixels samples in $200$ dimensional generic deep features, by orders of magnitudes. As long as the empirical error of decision trees is kept at the same level as a deep network, its expected error on unseen objects  will be significantly smaller.


\subsection{Postprocessing of Segmentation Masks}
Our initiative is to assign a label for each pixel by the confidence map SDForest prediction in continuous value. 
We use the concept of superpixel \cite{ren2003learning} and try to merge each large segment of pixels in irregular shapes to share uniform segmentation confidence. A segment could be  formed by grouping pixels by proximity, color, texture, and possibly other low-level details. and it preserves better geometry and boundary information comparing to pixels which are digitalization results. Typically this can be obtained by two families of algorithms, the graph-based \cite{shi2000normalized} or clustering-based ones \cite{achanta2012slic}. 
In our implementation, we use SLIC \cite{achanta2012slic} in this step. SLIC needs to compute the image gradients $\nabla I$ using neighboring pixel differences, and regularly select $k$ centers of superpixels  with uniform distance. The superpixel centers are then moving to the nearest pixels that have the smallest gradients. The indices of each pixel are then assigned by the nearest-neighbor principle, and the superpixel centers are updated accordingly. This procedure repeats itself several times until a certain criterion meets. After the clustering procedure we perform \textit{soft mean pooling} on the confidence level on each segment, which is average over the confidence over the same superpixel and assign the mean value as an offset to all pixels contained within.


 \begin{figure*}[htbp]
\centering 
\includegraphics[width=\textwidth]{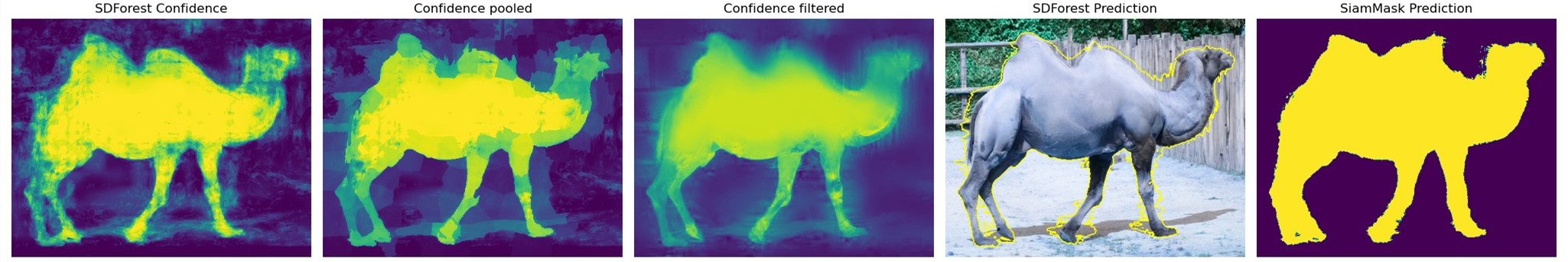}
\vspace{-10pt}
{\includegraphics[width=\textwidth]{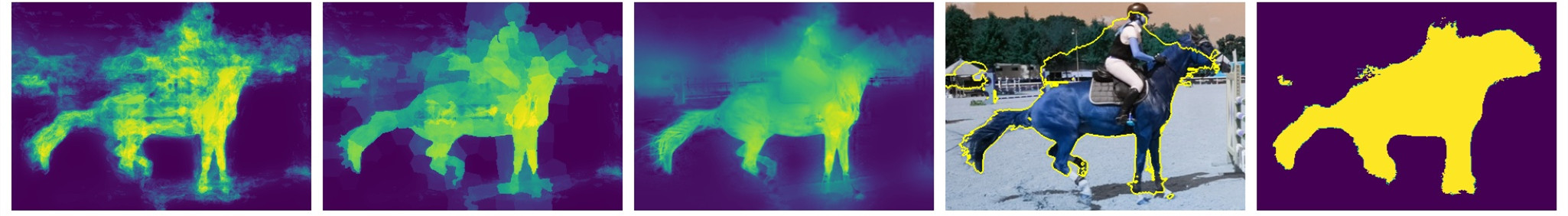}}
\vspace{-10pt}
{\includegraphics[width=\textwidth]{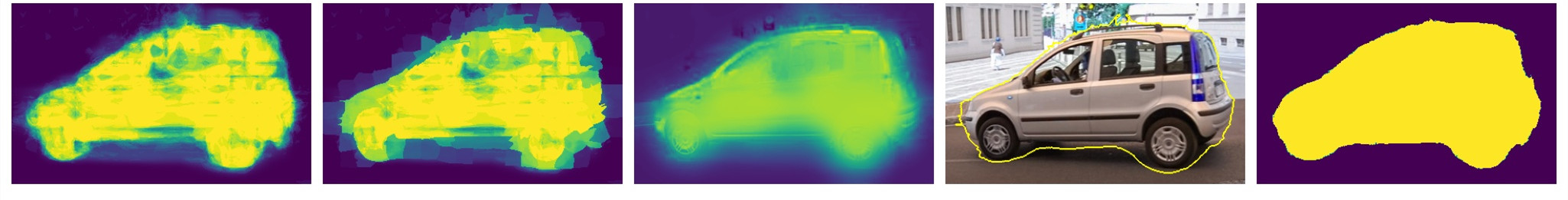}}
\vspace{-10pt}
{\includegraphics[width=\textwidth]{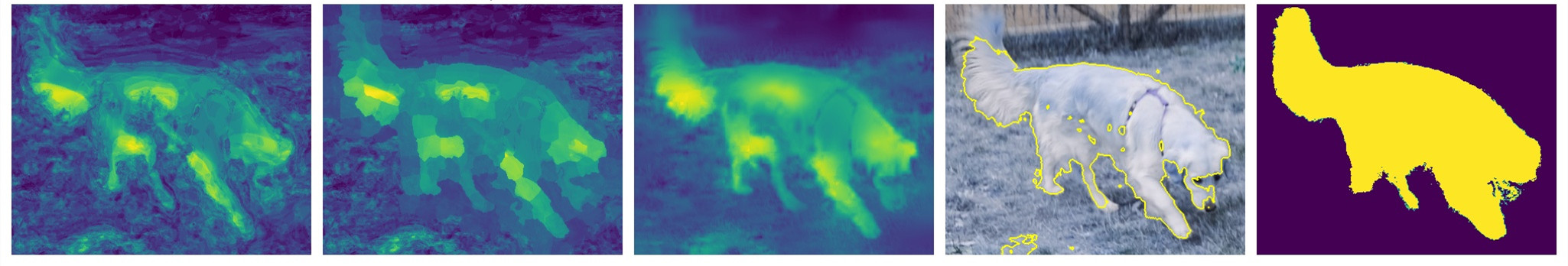}}
\vspace{-10pt}
{\includegraphics[width=\textwidth]{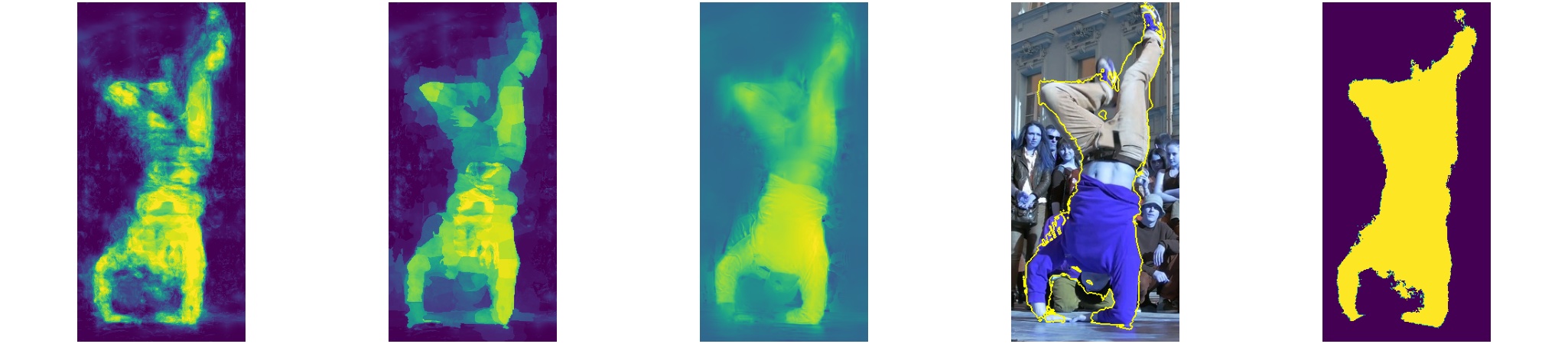}}
\vspace{-10pt}
\caption{ SDForest on Davis2016. From left to right: semiparametric forest regression, superpixel pooling, image guided filtering, confidence thresholding and the result from SiamMask. 
}
\label{fig:davis}
\vspace{-10pt}
\end{figure*}

The next step aims to provide a fine-grain tool to enhance the detail of the result, as previous steps are essentially statistical estimation and therefore lose attention to study how to round up the confidence of each pixel to zero or one, which ideally should be affected by nearby pixels as a joint prediction. We apply image-guided filter (IGF) \cite{he2010guided} to further adjust the border of confidence map and to increase the resolution to the full size of input images if SDForest applies on low-resolution deep features.  IGF is a linear filter based on a guidance image, in this case, the video frame in its the RGB representation $I$, and an input to be filtered, in this case, the confidence map $P(I)$. We continue to use a single index letter, e.g. $i,k$ to represent a $2D$ location. The output of the filter $Q$ should align its edges, corners and other low-level details with the raw image, 
\begin{align}
\nabla Q[i]= a \nabla I[i], \quad Q \in [0,1]^{w\times h}
\end{align}
where $\nabla$ represents the 2D spatial gradient. IGF  assumes the following form of linear relation between the guidance image $I$ and the filtered map $Q$,
\begin{align}
Q[i] =a_k I[i]+b_k, \forall i \in \omega_k,
\end{align}
where $\omega_k$ is a window centered at the pixel $k$. The filtering coefficients are determined via minimizing the following energy function within the window
\begin{align}
\min E(a_k,b_k) =\sum_{i \in \omega_k} ((a_k I[i]+b_k-P[i])^2 + \epsilon a_k^2).
\end{align}
where the term $\epsilon$ is introduced to regularize the filtering coefficients $a$ from being too large. We denote the mean value and variance of $I$ within each window as $Ave$ and $Var$ respectively, then will be a closed-form solution as
\begin{align}
&a_k = \frac{\sum_{i \in \omega_k} I[i] P[i] - Ave(\omega_k) \bar P[k]}{Var(\omega_k)^2 +\epsilon},\\
&b_k = \bar P[k]- a_k \cdot Ave(\omega_k), \quad \bar P[k] =\frac{1}{\omega_k} P[i].
\end{align}
We define the average filter coefficients as 
\begin{align}
\bar a[i] =\frac{1}{|\omega|}\sum_{k \in \omega_i}a_k,\quad \bar b[i] =\frac{1}{|\omega|}\sum_{k \in \omega_i}b_k,
\end{align}
and we get the filtered segmentation confidence  $Q$. 
Intuitively the gradients $\nabla Q$ will be much smaller than $\nabla I$ as the coefficients $\bar a_i$ is a low-pass filtered version of $a_i$. After this step, we threshold the confidence to get the mask,
\begin{align}
   Q[i] = \bar a_i I[i]+\bar b_i,\quad  \hat y[i]=\mathbb I(Q[i] \geq \epsilon) \in \{0,1\}.
\end{align}

\subsection{Learning-Theoretical Perspectives}
Our key observation here is how to define the generalization for video object segmentation and tracking more cleverly, which is different from that of classification in learning theory literature, where training images and testing images are assumed to be drawn from the same distribution. We first study the generalization error caused by the post-processing technique. As they are solely based on the testing images themselves and the algorithm is nonparametric, which carries a sharply smaller VC dimension or model complexity. 
In the following, we use a simplified analogy to analyze the generalization error by inter-pixel adjustment. 

We denote the pixel features $\{x[i]\}_{i=1}^m$ from the space $ \mathcal D$ and pixel labels $\{y[i]\}$.  Suppose the regressor $f(x,y)=w^{\top} \phi(x,y)$, where $f \in \mathcal F$, $y \in \{0,1\}^m$, $\phi(x,y)$ are potential functions that include unary and high-order relations. The prediction is by the rule of $\hat y=\arg\max_y f(x,y)$. Suppose that the loss function per-pixel is bounded by $C$ and $\|w\|\leq B$ for some constant $B,C$, and the expected max-margin loss $\mathcal L_D$ has a bound as (\cite{mohri2018foundations}, Theorem 13.2), for any $\delta \in (0,1)$, with a probability of at least $1-\delta$, there is
\begin{align*}
\mathcal L_D(f) \leq & (\log Z_w)/m + 8 B\cdot C \cdot \mathcal R_m(\mathcal F) \\
&+\sqrt{\log \log (4 B)}/\sqrt{m} + \sqrt{\log 2/\delta}/\sqrt{2m},
\end{align*}
where $\mathcal R_m(\mathcal F)$ is the empirical Rademacher complexity of the hypotheses set $\mathcal F$ on $m$ samples, and the normalizer $Z_w =\prod_i \sum_{y \in \mathcal Y} \exp(f(x_i,y) - f(x_i,y_i))$ depends on the training set of learning-based methods. Intuitively, the third term in R.H.S. characterizes convergence w.r.t. the number of pixels and the second term indicates that a classifier with smaller weights generalizes better. For a learning-based algorithm that needs training sets, $w$ is necessarily large to cover complex inter-pixel relations within the range of one window or one superpixel, which results in a larger generalization error. 


We then put the whole pipeline into learning theoretical analysis. For convenience, we will refer to our framework that transfer existing features and instantly learn a simple model from the first frame as instant learning. There are several advantages of instant learning for analysis. 1), A deep neural network naturally consists of cascade regression component plus nonlinear activation and thus is considered as of high model complexity, but once we use generic deep features from irrelevant tasks, they are treated as higher-dimensional features, which scales linearly to provide richer information, instead of adding exponentially large model complexity; 2), As training are only needed on the first frames, the rest frames, even with labels, serve as verification data. In contrast, a fully learning-based algorithm tend to overfit on the test data as the benchmark provide their corresponding labels as well, and thus has a perform gap when facing real-world data. We provide a comparison in Table.(1) to show how we rearrange the loss functions to more the generalization more strict.

\begin{table*}[h]
\label{table:theory}%
\caption{A learning theoretical characterization: how to assign the segmentation error on different images into the loss function.}
\vspace{-10pt}
\begin{center}
\begin{tabular}{c|c  c |c c c|}
\hline & \text{In datasets:} $I^k_1$ & $I^k_{2,\cdots,N}$  & \text{Unseen videos:} $I^0_1$ &$I^0_{2,\cdots,N}$ \\
\hline
\text{End-to-end learning} &$\mathcal L^{train}(f,h)$ & $\mathcal L^{train}(f,h)$ & $\mathcal L^{train}(f,h)$ &$\mathcal L^{test}(f,h)$  \\
\text{Instant learning}  & $\mathcal L^{train}(f)$  & $\mathcal L^{verify}(f)$ & $\mathcal L^{train}(f)$ &$\mathcal L^{test}(f)$   \\
\hline
\end{tabular}
\vspace{-10pt}
\end{center}
\end{table*}

In this case, we view each video sequence, indexed by $k=1,\cdots, K$, are for $K$ different tasks. We denote $\mathcal I =\{I_1^k\}$ as the training sets, and use $ I_1^k$ to denote the first frame of $k$th video. We define $F=\{f_k\}^k$ and $f_0$ as as the model for seen and unseen objects. Our instant learning framework uses a shared deep network $h$ for feature extraction and individual models $f_k$ for each video object, and our learning objective is to: for the seen objects in datasets, to minimize the training error on the first frame, use the rest frames as verification, and the test error will be measured on unseen objects $k=0$.
\begin{align}
\hat f_k= \arg\min_{f \in \mathcal F}\mathcal L^{train}(f)=  \ell(f_k \cdot h(I_1^k),y_1),\nonumber\\
\mathcal L^{verify}(f )=\frac{1}{K N} \sum_{k=1}^K  \sum_{n=2}^N \ell(f_k \cdot h(I_1^k),y_1)
\end{align}
By comparison, an end-to-end deep learning approach jointly optimize the feature function and the classifier $f$,
\begin{align}
\hat f,\hat h =\arg\min_{f \in \mathcal F} \min_{h \in \mathcal H}   \frac{1}{KN} \sum_{k=1}^K  \sum_{n=1}^N \ell(f_k \cdot h(I_n^k),y_n),
\end{align}
where the cascading of $h$ and $f$ multiplies the VC dimensions together according to Eq.(\ref{vc-dl}). 

We also try to explain that learning individual model $f_k$ for each video that generates diversified models will generalize better. 
and we follow the work in a theory of transfer learning \cite{tripuraneni2020theory}, to define the empirical difference $\bar d$ and worst-case difference $d$ of the joint feature extractor $h$ as the
\begin{align}
&\bar d_{F,f_0}(h';h) =  \min_{f \in \mathcal F} \mathbb E_{I}\left\{\ell(f'\cdot h'(x_j),y_j) -\ell(f'\cdot h(x_j),y_j) \right\},\nonumber\\
& d_{\mathcal F,f_0}(h';h) = \max_{f_0 \in \mathcal F} \bar d_{F,f_0}(h';h),
\end{align}
which characterizes how significantly would different representations $h$ change the empirical error. 
The segmentation function $F$ on video sequences are defined to be $(\nu,\epsilon)$-diverse over $\mathcal F$ and a given feature extractor $h$, if for all $h' \in \mathcal H$, there is
\begin{align}
\bar d_{ F,f_0}(h';h) \leq d_{\mathcal F,f}(h';h)/\nu +\epsilon.
\end{align}
Suppose that $f\in \mathcal F$ is $L$-Lipchitz, the loss function $\ell$ is bounded by $C$ and $\|h(I)[i]\| \leq D, \forall i$, and we denote $\mathcal G$ to be the Gaussian complexity of a model. By the definitions introduced above, with a probability of at least $1-2\delta$, the generalization error could be estimated by the following bound,
\begin{align*}
\mathcal L_D(f_0) &\leq \mathcal L_S(f_0)+ O \bigg (L^2 \log (K) /\nu+ L \cdot \mathcal G(F)\\  &+C\frac{1}{\nu}\sqrt{\frac{\log (2/\delta)}{K}}+\sqrt{\frac{\log (2/\delta)}{m}}
+\frac{LD}{\nu K^2} \bigg)
\end{align*}
This gives a quantitative explanation to an intuitive phenomenon, why the diversity in datasets contribute to better generalization, even they are merely used as verification and without providing information to fine-tune the models.
\begin{table*}[h]
\caption{Performance comparison on DAVIS 2016. FT represents whether it is fine tuned on the DAVIS dataset and M  represents whether it needs a mask in reference frame.}
\label{table:davis}
\begin{center}
\begin{tabular}{c|c c | c c c| c c c |c c}
\hline
& FT &M &$\jmean$ &$\jrecall$ &$\jdecay$ &$\fmean$ &$\frecall$ &$\fdecay$ &\text{speed}\\
\hline
MSK \cite{perazzi2017learning} &$\checkmark$ &$\checkmark$  & 79.7  &  93.1 &  8.9  & 75.4 &  87.1  & 9.0  & 0.1 \\
SFL \cite{cheng2017segflow} &$\checkmark$ &$\checkmark$  & 76.1  & 90.6  & 12.1  & 76.0  & 85.5  & 10.4 &  0.1 \\
FAVOS \cite{cheng2018fast} &$\xmark$ &$\checkmark$  & 82.4 &  96.5  & 4.5  & 79.5  & 89.4  & 5.5  & 0.8\\
RGMP \cite{wug2018fast} &$\xmark$ &$\checkmark$  & 81.5  & 91.7 &  10.9 &  82.0  & 90.8  & 10.1 &  8 \\
PML \cite{chen2018blazingly} &$\xmark$ &$\checkmark$ &  75.5 &  89.6 &  8.5  & 79.3  & 93.4  & 7.8  & 3.6 \\
OSMN \cite{yang2018efficient} &$\xmark$ &$\checkmark$  & 74.0  & 87.6  & 9.0  & 72.9  & 84.0  & 10.6  & 8.0 \\
PLM \cite{shin2017pixel} &$\xmark$ &$\checkmark$  & 70.2  & 86.3  & 11.2  & 62.5  & 73.2  & 14.7 &  6.7 \\
VPN \cite{jampani2017video} &$\xmark$ &$\checkmark$  & 70.2  & 82.3  & 12.4  & 65.5  & 69.0  & 14.4  & 1.6 \\
SiamMask \cite{wang2019fast} &$\xmark$ &$\xmark$  & 71.7  & 86.8  & 3.0  & 67.8 &  79.8  & 2.1  & 55\\
\hline
SDForest &$\xmark$ &$\xmark$  & 73.8  &  87.6   & 5.8  & 68.9  &  78.8   & 7.8 &45 \\
\hline
\end{tabular}
\vspace{-10pt}
\end{center}
\end{table*}
\begin{table*}[h]
\caption{Performance comparison on DAVIS 2017. FT represents whether it is fine tuned on the DAVIS dataset and M  represents whether it needs a mask in reference frame.}\label{table:davis2017}
\begin{center}
\begin{tabular}{c|c c | c c c| c c c |c}
  & FT &M &$\jmean$ &$\jrecall$ &$\jdecay$ &$\fmean$ &$\frecall$ &$\fdecay$ &\text{speed}\\
\hline
OnAVOS \cite{voigtlaender2017online} &$\checkmark$& $\checkmark$ & 61.6  & 67.4  & 27.9  & 69.1  & 75.4  & 26.6  & 0.1 \\
OSVOS \cite{caelles2017one}  & $\checkmark$ & $\checkmark$ & 56.6  & 63.8  & 26.1  & 63.9  & 73.8  & 27.0  & 0.1 \\
FAVOS \cite{cheng2018fast}  & $\xmark$  & $\checkmark$ & 54.6  & 61.1  & 14.1  & 61.8  & 72.3  & 18.0 &0.8 \\
OSMN \cite{yang2018efficient}  & $\xmark$  & $\checkmark$ & 52.5  & 60.9  & 21.5  & 57.1  & 66.1  & 24.3  & 8.0 \\
SiamMask \cite{wang2019fast}  & $\xmark$  & $\xmark$  & 54.3  & 62.8  & 19.3  & 58.5  & 67.5  & 20.9  & 55 \\
\hline
SDForest & $\xmark$  & $\xmark$  & 55.2  &  64.5   &  21.4  & 59.0  &  68.8   &  23.4  & 45\\
\hline
\end{tabular}
\vspace{-10pt}
\end{center}
\end{table*}
\section{Experiments}
We conduct the experimental study on DAVIS 2016 challenge dataset \cite{perazzi2016benchmark},  a high-quality video dataset that includes segmentation masks for each frame. Following the original paper of the dataset and other works in this area, we borrow the object tracking part of SiamMask\cite{wang2019fast} in their open-sourced implementation, to track the object in their coarse positions to focus on such a search window, which certainly could be changed to other even faster trackers. We will mainly compare SDForest with SiamMask, as it is a real-time deep learning algorithm that needs no additional training on the datasets, granted that several other works either have slower speed or additional fine-tuning achieve better accuracy. 
For generating features, we choose to use a recent work,  EfficientNet-B$0$ \cite{tan2019efficientnet} which is trained on ImageNet. This network is the fastest one in its family, and we interpolate the layers of depth $1,5,9,13$ to the resolution of the search window, which gives a concatenation of $219$ channels in the original resolution. The network structure of SiamMask is based on the backbone of ResNet-50. According to the training procedure described in the original paper, the network was pre-trained on ImageNet-1\textit{k} dataset, and then trained on Microsoft COCO using an SGD optimizer with $20$ epochs. The sizes of the examplar patch and the search window are $127\times 127$ and $256\times 256$, respectively.  The confidence map branch and the refinement branch are both built upon this tensor. We use an implementation of a random forest classifier from \textit{Scikit-Learn}.  For each video, we process the first frame  and construct the forest estimator. The forest has $20$ trees of maximum depth $20$, they are split by the maximum Gini increase principle. We give an example of one depth-$2$ tree in Figure.(\ref{fig:tree}). 
We construct the training pixels from the initial frame as follows: 1) pixels within the search windows are all included, along with their segmentation labels. 2) pixels outside of the search windows are included at a stride of $s=10$ horizontally as negative examples. 
This design ensures us to sample a large number of positive pixels within the box, along with sufficient negative pixels across the image, so that the forest classifier would distinguish the target with hard negative pixels in the distance which the bound box may encounter in the following frames. e.g. in the swan example in Figure.(\ref{fig:workflow}), where a reflection of the swan in the lake resembles its appearance and may mix with the real target, see the right-bottom area of Figure.(\ref{fig:sfig1}). It is worth mentioning that SiamMask features are not suitable for forest since it uses a large area of zero paddings.  On top of deep features, we denote the forest classifier prediction as $\mu(I) \in [0,1]^{w \times h}$, and we construct a logistic regression estimator $\tilde P_{\theta}(I)\in [0,1]^{w \times h}$ on the same training pixels with L-BFGS, and generate their ensemble model $0.8P_{\theta}(I) +0.2 \tilde P_{\theta}(I)$. We plot the confidence map in Figure.(\ref{fig:sfig1}), which shows that the semi-parametric model outperforms others. 


We use two measures, F-measure ($\mathcal F$) and J-measure ($\mathcal J$), which counts for pixel-wise similarity and contour-wise similarity. There are three kinds of statistics for each measure, namely, mean $\jmean,\fmean$, recall  $\jrecall,\frecall$, and decay $\jdecay,\fdecay$. We show the result in Table.(\ref{table:davis}) and Figure.(\ref{fig:davis}).  In Figure.(\ref{fig:davis}) we see that SDForest generates considerably detailed maps, e.g. the legs of camel, horse, dogs and humans are even clearly separated. We highlight the efficiency of SDForest and SiamMask as both algorithms use lightweight deep networks, and not use any data from DAVIS 2016. Both models are incredibly efficient in inference, generating real-time segmentation, which most contemporaries cannot achieve, despite that they might have better accuracy.  To give expression on the efficiency: constructing random forest on the reference of DAVIS 2016 \textit{SWAN} example, $82,335$ selected pixels with $219$-dimensional features, takes $3.27$ seconds for $50$ consecutive frames, averaging to about $60$ ms per frame, on laptop level GPU and CPU, comparing to $1.42$ seconds for a single forward-propagation of EfficientNet on the whole image in one frame. Our comparison with other recent works demonstrates that our proposed method improves upon SiamMask with a considerable gap. Although there cannot be visually prominent changes over SiamMask, restricted by SDForest being extremely light, still, some corner and edge details of segmentation, like the tail of swans and the back of camels are clearly tighter.
We perform experiments on DAVIS 2017 for multiple object segmentation task, with the result in Table.(\ref{table:davis2017}). The implementation is mostly the same as that of DAVIS 2016, except for that for multiple object instance segmentation, we need to predict each pixel out of confidence maps of multiple objects. We would highlight that SDForest is relatively more efficient when dealing with multiple objects, as it only needs one forward propagation on the whole frame and gets the confidence map for all objects, each of which needs one round of forward propagation for deep network predictions. In the table, we see that some stronger competitors are showing better performance but with a much slower speed, considering both efficiency and independence on training data, SDForest consistently outperforms several competitors including SiamMask, which has better efficiency due to integration of segmentation header and tracker in one network.
\section{Conclusion}
While the research community has been promoting advancement in object segmentation with over-parameterization and end-to-end learning, we try to prove that a simple prompting module can adapt foundational models to specific tasks, maintaining accuracy and increase generalization capability on deep features. Though the approaches cannot be included in one work, we believe this non-mainstream thinking paradigm to be beneficial for developing practical, economical, and reliable algorithms.

{\small
\bibliographystyle{ieee_fullname}
\bibliography{cvpr}
}

\end{document}